\pdfoutput=1

\documentclass[11pt]{article}

\usepackage[table]{xcolor}
\usepackage{array}
\usepackage[final]{acl}

\usepackage{times}
\usepackage{latexsym}

\usepackage[T1]{fontenc}

\usepackage[utf8]{inputenc}

\usepackage{microtype}

\usepackage{inconsolata}

\usepackage{graphicx}

\usepackage{booktabs}
\usepackage{multirow}
\usepackage{enumitem}

\usepackage{arydshln}
\usepackage{hhline}
\usepackage{placeins}
\usepackage{tcolorbox}
\usepackage{soul}

\usepackage{bm}
\usepackage{amsmath}
\usepackage{amsfonts}

\usepackage{tikz}
\usetikzlibrary{shapes.geometric}

\definecolor{neuron}{HTML}{377EB8}
\definecolor{sae}{HTML}{FF7F00}
\definecolor{attention}{HTML}{F781BF}
\definecolor{circuit}{HTML}{4DAF4A}

\definecolor{evalPred}{rgb}{0.6, 0.4, 0.8}   
\definecolor{evalInput}{rgb}{0.2, 0.6, 0.6}   
\definecolor{evalOutput}{rgb}{0.85, 0.3, 0.3}
\definecolor{evalSem}{rgb}{0.45, 0.5, 0.6}    
\definecolor{evalHuman}{rgb}{0.8, 0.5, 0.2}   

\DeclareRobustCommand{\methodtag}[2]{%
  \tikz[baseline=(char.base)]{%
    \node[
      fill=#1,
      rounded corners=2pt,
      text=white,
      font=\bfseries,
      minimum width=1em,
      minimum height=1em,
      align=center,
      inner sep=0pt
    ](char){#2};%
  }%
}

\DeclareRobustCommand{\evaltag}[2]{%
  \!
  \texorpdfstring{%
  \tikz[baseline=(char.base)]{%
    \node[
      shape=diamond,
      fill=#1,
      text=white,
      font=\bfseries,
      minimum size=1.5em, 
      align=center,
      inner sep=0pt
    ](char){#2};%
  }%
  }
}

\newcommand{\neurontag}{\protect\methodtag{neuron}{N}}
\newcommand{\saetag}{\protect\methodtag{sae}{S}}
\newcommand{\attentiontag}{\protect\methodtag{attention}{A}}
\newcommand{\circuittag}{\protect\methodtag{circuit}{C}}

\newcommand{\predtag}{\protect\evaltag{evalPred}{P}}
\newcommand{\inputtag}{\protect\evaltag{evalInput}{I}}
\newcommand{\outputtag}{\protect\evaltag{evalOutput}{O}}
\newcommand{\semtag}{\protect\evaltag{evalSem}{$\approx$}}
\newcommand{\humantag}{\protect\evaltag{evalHuman}{H}}

\title{
    Interpreting Language Models Through Concept Descriptions: A Survey
}

\newcommand{\affilsup}[1]{\rlap{\textsuperscript{\normalfont#1}}}

\author{
    Nils Feldhus*\affilsup{1,2}
    \qquad
    Laura Kopf*\affilsup{1,2}
    \\
    $^1$BIFOLD – Berlin Institute for the Foundations of Learning and Data \\
    $^2$Technische Universit\"at Berlin \\
    {\small \texttt{\{feldhus,kopf\}@tu-berlin.de}} \\
    {\footnotesize *Equal contribution}
}

\begin{document}
\maketitle
\begin{abstract}
    Understanding the decision-making processes of neural networks is a central goal of mechanistic interpretability. In the context of Large Language Models (LLMs), this involves uncovering the underlying mechanisms and identifying the roles of individual model components such as neurons and attention heads, as well as model abstractions such as the learned sparse features extracted by Sparse Autoencoders (SAEs). 
    A rapidly growing line of work tackles this challenge by using powerful generator models to produce open-vocabulary, natural language concept descriptions for these components.
    In this paper, we provide the first survey of the emerging field of concept descriptions for model components and abstractions. We chart the key methods for generating these descriptions, the evolving landscape of automated and human metrics for evaluating them, and the datasets that underpin this research.
    Our synthesis reveals a growing demand for more rigorous, causal evaluation.
    By outlining the state of the art and identifying key challenges, this survey provides a roadmap for future research toward making models more transparent.
\end{abstract}

\section{Introduction}

The interpretability of large generative models, and specifically LLMs, poses a central challenge to understanding their internal computations and enabling transparent decision-making.
A key goal of this effort, often termed mechanistic interpretability, is to reverse-engineer the algorithms learned by these models through analysis of their fundamental components, such as individual neurons and attention heads \cite{saphra-wiegreffe-2024-mechanistic, ferrando-2024-primer-inner-workings-transformers}.
A central problem in component analysis is assigning human-understandable meaning to these building blocks. Early approaches sought to map component activations to predefined linguistic properties through probing classifiers \cite{conneau-2018-cram, belinkov-2021-probing} or to test their alignment with human-defined concepts \cite{kim-2018-tcav, lee-2024-from-neural-activations-to-concepts}. While foundational, these methods are limited by their reliance on a fixed set of concepts, potentially missing the novel representations learned by the model itself.

\begin{figure}[t]
    \centering
    \resizebox{.85\columnwidth}{!}{%
        \includegraphics{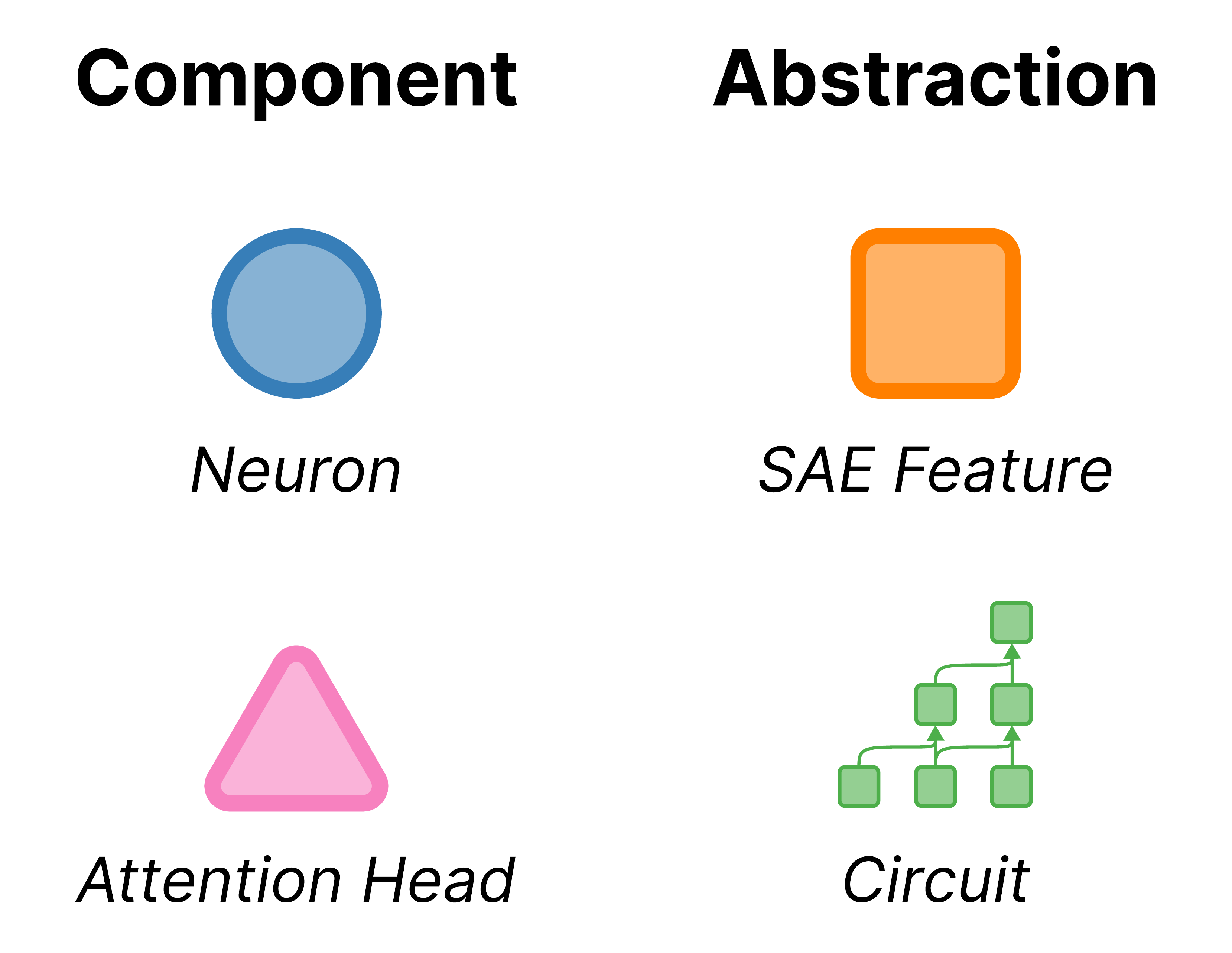}
    }
    \caption{Illustration of model components (left column) and model abstractions (right column) of a language model. Components are individual units of a language model, such as neurons or attention heads. Abstractions go beyond the individual components of a model encompassing higher-level representations such as those learned by SAEs, or subgraphs involving multiple components or sparse features, as in circuits. Each component or abstraction can be associated with a human-understandable concept description.}
    \label{fig:overview}
\end{figure}

A recent paradigm shift leverages the generative power of LLMs to overcome this limitation. Instead of testing for predefined concepts, this new line of work uses LLMs to produce open-vocabulary concept descriptions for the components of another model. This approach elicits a natural language explanation of a component's function by prompting the generator with data on when that component activates \cite{bills-2023-explain-neurons, cunningham-2024-saes-find-highly-interpretable-features, choi-2024-scaling-automatic-neuron-description}. 
For instance, given text fragments that maximally activate a specific neuron, an LLM synthesizes a description of the concept that neuron appears to detect, such as ``legal clauses'' or ``references to the 1980s''. 
This method can be applied to both native model components and learned abstractions like Sparse Autoencoder (SAE) features (Figure~\ref{fig:overview}).

This survey provides a structured overview of this rapidly emerging field. We focus on the methods, datasets, and evaluation techniques for generating concept descriptions for LLM internals, and we aim to answer the following questions:
\begin{enumerate}[topsep=1pt, partopsep=0pt, itemsep=0pt, leftmargin=0.7cm]
\renewcommand{\labelenumi}{(\theenumi)}
    \item What are common methods and data sources for generating concept descriptions for interpreting language models? (\S \ref{sec:description-methods})
    \item What are best practices for evaluating such concept descriptions? (\S \ref{sec:evaluation})
    \item What are the common trends that can be observed in concept description research? (\S \ref{sec:findings})
    \item What are the key gaps and promising future directions for concept-based descriptions of language models? (\S \ref{sec:future-work})
\end{enumerate}

\section{Definitions}
\label{sec:definitions}

\subsection{Language Models}

\paragraph{Neuron}
At the heart of modern LLMs lies the Transformer architecture \cite{vaswani-2017-attention}, which stacks layers composed of two primary sub-modules: multi-head self-attention (MHA) and a position-wise feed-forward network (FFN). While MHA computes context-aware token representations, the FFN is responsible for further transforming these representations. An FFN layer typically consists of two linear transformations with a non-linear activation function in between:
\begin{equation}
    \bm{h}^i = \text{act}_{\text{fn}}(\tilde{\bm{h}}^i\bm{W}^i_1) \cdot \bm{W}^i_2,
\end{equation}
where $\tilde{\bm{h}}^i$ is the output from the attention sub-layer, and $\bm{W}^i_1 \in \mathbb{R}^{d \times d_{ff}}$ and $\bm{W}^i_2 \in \mathbb{R}^{d_{ff} \times d}$ are weights matrices ($d_{ff}$ is typically $4d$).

Within this framework, a neuron is defined as one of the intermediate dimensions in the FFN; specifically, it corresponds to a single column in the first weight matrix $\bm{W}^i_1$ and its subsequent non-linear activation \cite{geva-2022-ffn-promoting-concepts,sajjad-2022-survey-neuron-level-interpretation}. A neuron is considered to have \textit{fired} or \textit{activated} for a given input if its activation value, $\text{act}_{\text{fn}}(\tilde{\bm{h}}^i\bm{W}^i_1)_j$, is positive.

However, the neuron as a unit of interpretation is fundamentally challenged by the phenomenon of polysemanticity. A single neuron often activates for a diverse and seemingly unrelated set of inputs, making its function difficult to summarize with a single, coherent description. This is a consequence of superposition, a strategy where models represent a vast number of concepts by encoding them as linear combinations of neuron activations \cite{elhage-2022-superposition}. While this is an efficient use of limited parameters, it means that individual neurons are inherently entangled and do not typically correspond to clean, understandable concepts.

\paragraph{Attention Head}
A level up from the individual neuron is the attention head. As a core component of the Transformer architecture, the multi-head attention mechanism allows a model to process information from different representation subspaces in parallel. Each head can be conceptualized as a specialist that learns to focus on different parts of the input sequence to capture specific relational patterns between tokens.

A significant body of research has demonstrated that these heads often acquire specialized and interpretable roles. For example, \citet{voita-2019-analyzing-multi-head-self-attention} categorized attention heads in a machine translation model into distinct functional groups, including heads that attend to adjacent tokens (positional), heads that implement syntactic dependencies, and heads that focus on rare words. Subsequent work has identified heads responsible for more complex linguistic phenomena like subject-verb agreement, coreference resolution \cite{clark-2019-what-does-bert-look-at}, and dependency parsing \cite{shen-2022-unsupervised-dependency-graph-network}, as well as newer paradigms such as in-context retrieval augmentation \cite{kahardipraja-2025-atlas-in-context-learning}.

\subsection{Concepts}

In the context of LLM interpretability, concepts constitute a human-understandable description of a pattern or property that a model has learned to represent. 
This notion aligns with the definition in \citet{dalvi-2022-latent-concepts-in-bert} who describe concepts as meaningful groups of words that can be clustered by a shared linguistic relationship.
Such relationships can span a wide spectrum of abstraction, from low-level lexical features (e.g., \textit{words starting \mbox{with ``anti-''}}) and syntactic roles (e.g., \textit{direct objects}) to high-level semantic categories (e.g., \textit{names of capital cities}, \textit{legal terminology}, or \textit{financial terms}).

In terms of selecting ground-truth concepts, many works rely on pre-defined concepts, e.g., by using metaclasses in Wikipedia \cite{schwettmann-2023-find,dumas-2025-separating-tongue-from-thought}, or concept classification datasets \cite{antognini-faltings-2021-conrat,abraham-2022-cebab,jourdan-2023-cockatiel,singh-2023-sasc,sun-2025-eco-concept}.
Others manually determine ground-truth concept labels with the help of human annotators \cite{dalvi-2022-latent-concepts-in-bert,mousi-2023-can-llms-facilitate-interpretation} or validate concept explanations post-hoc, e.g., in terms of usefulness for understanding the model's classification decision in a downstream task \cite{yu-2024-lacoat}.
Recent advances for topic modeling involving LLMs include SEAL \cite{rajani-2022-seal}, a labeling tool for identifying challenging subsets in data and assigning human-understandable semantics to them.
LLooM~\cite{lam-2024-lloom}, which extracts interpretable, high-level concepts from unstructured text. It uses model embeddings with clustering methods and is shown to be better aligned with human judgment of semantic similarity and topic quality. 
Goal-driven explainable clustering~\cite{wang-2023-goalex} assigns free-text explanations to each cluster with an LLM. Such approaches focus on providing a thematic summarization of textual inputs.

\subsection{Natural Language Explanations}
The automatic generation of human-readable text has been a long-standing goal in NLP. Natural Language Generation (NLG) systems often relied on templates and hand-crafted rule sets to convert structured data into prose \cite{gatt-2018-survey-nlg}. This landscape shifted dramatically going via sequence-to-sequence architectures to LLMs and the prompting paradigm, also effecting how to generate explanations:
Instead of simply predicting extractive rationales \cite{lei-2016-rationalizing}, a natural language explanation (NLE) explains model predictions with free text \cite{camburu-2018-esnli,wiegreffe-2022-reframing}. This gives them greater expressive power in terms of the reasoning they can convey, especially with complex reasoning tasks necessitating implicit knowledge.

\begin{figure*}[t]
  \begin{center}
    \centerline{\includegraphics[width=\textwidth]{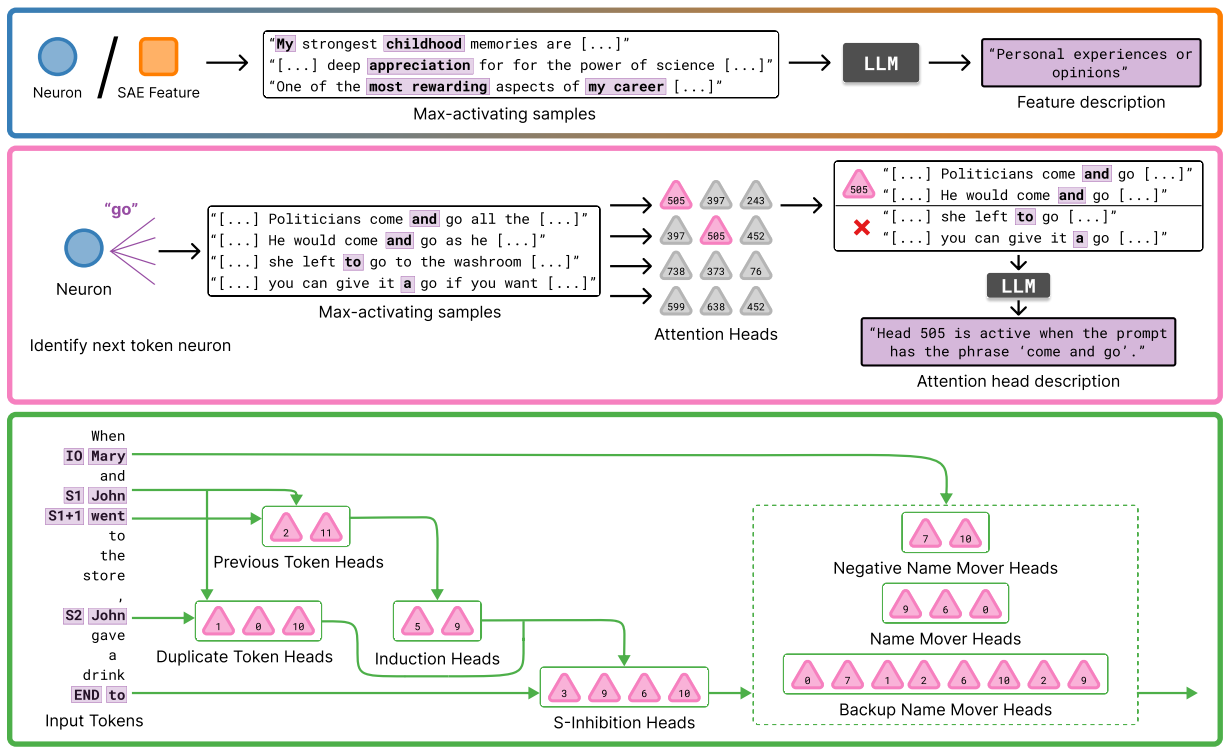}}
    \caption{Overview of descriptions for model components (\neurontag{} neurons, \attentiontag{} attention heads) and model abstractions (\saetag{} SAE features, \circuittag{} circuits). The top panel shows a schematic example of an automatically generated feature description for a neuron or SAE feature, based on top-activating text samples (same process for both). The middle panel shows an example from \citet{neo-2024-interpreting-context-look-ups} of how attention head descriptions are generated: first a token-predicting neuron is identified, then prompts that highly activate it are found, the attention heads responsible for its activation are determined, and explanations for these heads are generated. The bottom panel shows a circuit from \citet{wang2023interpretability} implementing indirect object identification (IOI), where input tokens enter the residual stream and attention heads move information between streams, query/output arrows indicate where they write, key/value arrows where they read, and each class of head has an associated description.}
    \label{fig:examples}
  \end{center}
  \vspace*{-1em}
\end{figure*}

NLEs generated by LLMs exceed other explanation methods in plausibility \cite{jacovi-goldberg-2020-towards}, but lack in faithfulness guarantees, as evidenced by a slew of recent studies \cite{turpin-2023-unfaithful,lanham-2023-measuring-faithfulness,chen-2023-cf-sim,madsen-2024-are-self-explanations-faithful,bentham-2024-cot-unfaithfulness,parcalabescu-frank-2024-cc-shap,bartsch-2023-self-consistency}.
Furthermore, the resulting descriptions are highly sensitive to the specific prompt and the choice of the generator model, so the same prompt can yield different explanations. As noted by \citet{bills-2023-explain-neurons}, LLMs often produce overly broad or generic summaries and do not capture the nuance of the specific concept(s) represented by a model component. Figure \ref{fig:examples} presents examples of such descriptions for different model components and abstractions.

\subsection{Concept Datasets}
The concepts a language model learns are fundamentally constrained by the data on which it was trained. Any dataset, from large-scale web scrapes to curated classification benchmarks, is an implicit repository of concepts. The frequency, context, and co-occurrence of words and phrases in the training corpus determine which patterns the model learns to represent internally. Consequently, interpretability methods that aim to describe internal components are, in effect, attempting to reverse-engineer and label these data-driven concepts. This process inherently creates an interpretability illusion: A model likely cannot represent a concept that is absent from its training data and the descriptions are therefore bounded by the conceptual scope of the underlying dataset \cite{bolukbasi2021interpretability}.

Popular examples for datasets with explicit, human-labeled concepts include CEBaB \cite{abraham-2022-cebab} and aspect-based sentiment datasets like BeerAdvocate \cite{mcauley-2012-learning-attitudes} and Hotel reviews \cite{wang-2010-latent-aspect-rating-analysis}. These resources are invaluable for controlled experiments where the goal is to see if a model's internal representations align with pre-defined, human-validated concepts. On the other hand, they heavily restrict the vocabulary in concept descriptions and certainly exclude the potential of detecting concepts that are not obvious to the human observer \cite{hewitt-2025-we-cant-understand-ai}.

\begin{table*}[t]
    \centering
    \resizebox{\textwidth}{!}{%
        \renewcommand*{\arraystretch}{1.2}
    	\begin{tabular}{l >{\columncolor{neuron!40}}l >{\columncolor{sae!40}}l >{\columncolor{circuit!40}}l >{\columncolor{attention!40}}l ll}
        \toprule
        \multirow{2}{*}{\textbf{Study}} & \multicolumn{4}{c}{\textbf{Explained Model}} & \textbf{Description} & \textbf{Target}  \\
        & \textbf{\neurontag{} Neurons} & \textbf{\saetag{} SAEs} & \textbf{\circuittag{} Circuits} & \textbf{\attentiontag{} Attention H.} & \textbf{Source} & \textbf{Dataset} \\
        \midrule
        \citet{bills-2023-explain-neurons} & GPT-2 XL & & & & GPT-4 & WebText \\
        
        \multirow{2}{*}{\textls[-25]{\citet{cunningham-2024-saes-find-highly-interpretable-features}}} & & Pythia-70M & & & \multirow{2}{*}{GPT-4} & \multirow{2}{*}{OpenWebText} \\
        & & Pythia-410M & & & & \\
        
        \multirow{2}{*}{\citet{paulo-2024-automatically-interpreting-millions}} 
            & & Llama-3.1 8B, & & & Claude Sonnet 3.5, & \multirow{2}{*}{RedPajama-v2} \\
            & & Gemma-2 9B & & & Llama-3.1 70B  \\

        \textls[-50]{\citet{rajamanoharan-2024-jumping-ahead}} 
            & & Gemma-2 9B & & & Gemini Flash & \textit{unspecified} \\
        
        \multirow{2}{*}{\citet{choi-2024-scaling-automatic-neuron-description}} 
            & Llama-3.1 8B & & & & \multirow{2}{*}{GPT-4o-mini} & LMSYS-Chat-1M, \\
            &  & & & & & FineWeb \\
        
        \citet{he-2024-llama-scope}
            & & Llama-3.1 8B & & & GPT-4o & SlimPajama \\
        
        \multirow{3}{*}{\citet{gur-arieh-2025-output-centric}} 
            & Gemma-2 2B, & Gemma Scope, & & \\
            & Llama-3.1 8B, & Llama Scope, & & & GPT-4o-mini & The Pile \\
            & GPT-2 small, & OpenAI SAE & & \\
        
        \multirow{2}{*}{\citet{kopf-2025-prism}}
            & GPT-2 XL, & GPT-2 small, & & & Gemini-1.5-Pro, & \multirow{2}{*}{C4} \\
            & Llama-3.1 8B & Gemma Scope & & & GPT-4o-mini \\
        
        \multirow{2}{*}{\citet{chen-2025-knowledge-microscope}} 
            & & Gemma-2 2B, & & & \multirow{2}{*}{GPT-4o-mini} & \multirow{2}{*}{PrivacyParaRel} \\
            & & Gemma-2 9B & & & & \\
        
        \citet{heap-2025-saes-can-interpret-randomly} 
            & & Pythia 70M – 7B & & & Llama-3.1 70B & RedPajama-v2 \\
        
        \citet{muhamed-2025-decoding-dark-matter} 
            & & Pythia 70M & & & Claude 3.5 Sonnet & arXivPhysics \\
        
        \multirow{3}{*}{\citet{movva-2025-hypothesaes}}
            & & & & & & Wiki, Bills, \\
            & & OpenAI SAE & & & GPT-4o & Headlines, \\
            & & & & & & Yelp, Congress \\

        \citet{wang2023interpretability} & & & GPT-2 small & & \textit{human} & IOI \\

        \multirow{2}{*}{\citet{marks2025sparse}} 
            & & & Pythia 70M SAE, & & \multirow{2}{*}{\textit{human}} & \multirow{2}{*}{Bias in Bios} \\
            & & & Gemma Scope & & & \\
        
        \multirow{2}{*}{\citet{elhelo-geva-2024-maps}} 
            & & & & Pythia 6.9B, & \multirow{2}{*}{GPT-4o} & \multirow{2}{*}{\textit{custom}} \\
            & & & & GPT-2 XL \\
        
        \multirow{3}{*}{\citet{neo-2024-interpreting-context-look-ups}} 
            & & & & GPT-2, &  \\        
            & & & & Pythia 160M, & GPT-4 & The Pile \\
            & & & & Pythia 1.4B \\
        \bottomrule
        \end{tabular}
    }
    \caption{Concept description techniques categorized by component/abstraction (\neurontag{} Neurons, \saetag{} SAEs, \circuittag{} Circuits, \attentiontag{} Attention Heads), description source, and target dataset.}
    \label{tab:survey}
\end{table*}

 \section{Description Methods}
\label{sec:description-methods}
This section surveys the main targets of natural language concept description methods. These methods aim to make language models more transparent to humans in terms of understanding individual roles and underlying mechanisms. We organize these targets into native model components (\S\ref{sec:components}) and learned model abstractions (\S\ref{sec:abstractions}). A comparative overview is provided in Table~\ref{tab:survey}.

\subsection{Model Components}
\label{sec:components}

\paragraph{\neurontag{} Neurons} A growing line of research focuses on fully automated descriptions of model components, such as neurons and their associated functions. In computer vision, early work explored neuron-level interpretability through visual concept alignment and feature visualization, generating text-based explanations and concept annotations to make individual neurons more human-interpretable~\cite{bau-2017-network-dissection,mu-andreas-2020-compositional-explanations,hernandez-2022-milan,oikarinen2023clipdissect,bykov2023invert,kopf-2024-cosy}.
For language models, the seminal auto-interpretability approach by \citet{bills-2023-explain-neurons} proposed labeling all neurons in GPT-2 XL. Their method uses GPT-4 to generate textual explanations from input samples that strongly activate each neuron. Since its introduction, this approach has been widely adopted and extended in follow-up work on automated neuron description~\cite{choi-2024-scaling-automatic-neuron-description,gur-arieh-2025-output-centric,kopf-2025-prism}.

\paragraph{\attentiontag{} Attention Heads}
Another potential target component is the attention head. The \textsc{MAPS} framework, introduced by \citet{elhelo-geva-2024-maps}, provides a powerful example for inferring a head's functionality directly from its parameters, by projecting the head's weight matrices into the vocabulary space to create a token-to-token mapping matrix. This reveals the transformations the head has learned to perform, which lets us automatically map how strongly a given head implements a predefined relation, such as the knowledge-based \textit{Country to capital} mapping or the linguistic \textit{Word to antonym} operation. \citet{neo-2024-interpreting-context-look-ups}, illustrated in Figure~\ref{fig:examples}, investigate attention-MLP interactions. For example, they identify heads that perform ``copying'' from previous tokens in the same context, and use an LLM to generate and validate hypotheses about these functions.

\subsection{Model Abstractions}
\label{sec:abstractions}

\paragraph{\saetag{} SAE Features}

Most of the neuron description methods adopt decomposition-based approaches that assign a single description to each neuron~\cite{bills-2023-explain-neurons,choi-2024-scaling-automatic-neuron-description,gur-arieh-2025-output-centric}, thereby limiting interpretability to the latent dimensions of the original model. However, this strategy struggles with neuron polysemanticity, so recent research has shifted toward learning more disentangled representations.
One prominent direction involves sparse coding using SAEs \mbox{\cite{bricken-2023-towards-monosemanticity,shu-2025-sae-survey}}, which decompose model activations into higher-dimensional, sparsely activated feature spaces. These representations enable capturing a wider range of more interpretable, potentially monosemantic concepts~\cite{bricken-2023-towards-monosemanticity,templeton-2024-scaling-monosemanticity,gao-2025-scaling-evaluating-saes}.
Auto-interpretability techniques initially developed for neurons~\cite{bills-2023-explain-neurons} have since been successfully extended to these learned SAE features~\cite{cunningham-2024-saes-find-highly-interpretable-features, bricken-2023-towards-monosemanticity,gao-2025-scaling-evaluating-saes,he-2024-llama-scope, mcgrath-2024-mapping-latent-space}. However, SAEs have recently been subject to critique, with negative results reported by \citet{smith-2025-negative-results-for-saes} showing their underperformance relative to linear probes.

\paragraph{\circuittag{} Circuits} 
A further step beyond studying individual components or SAE features is the analysis of circuits, which are computational subgraphs composed of multiple interacting components. The goal is to understand how the components interact to accomplish a specific task, and represent this interaction as a human-interpretable graph that captures the computation responsible for the task. \citet{wang2023interpretability} and \citet{ameisen-2025-circuit-tracing} use manual analysis and human labeling to trace and understand circuits, such as those responsible for indirect object identification. More recently, \citet{marks2025sparse} have pioneered methods to automatically discover and describe ``sparse feature circuits'', demonstrating how compositions of SAE features can implement complex logic, such as detecting gender bias. The well-studied IOI circuit has since been formalized as a benchmark task in the MIB suite \cite{mueller-2025-mib}, providing a standardized way to evaluate circuit discovery methods. However, the fully automated generation of natural language descriptions for arbitrary, higher-order circuits remains an open challenge.

\begin{table*}[t]
    \centering
    \renewcommand*{\arraystretch}{.75}
    \setlength{\tabcolsep}{18pt}
    \resizebox{.875\textwidth}{!}{%
    \begin{tabular}{lll}
        \toprule
        \textbf{Evaluation Type} & \textbf{Measure} & \textbf{Study} \\
        \midrule
        
        \multirow{13}{*}{\predtag{} Predictive Simulation} & \multirow{12}{*}{Simulator Correlation} 
        & \citet{bills-2023-explain-neurons} \\
        & & \citet{lee-oikarinen-2023-importance-prompt-tuning} \\
        & & \citet{cunningham-2024-saes-find-highly-interpretable-features} \\
        & & \citet{he-2024-llama-scope} \\
        & & \citet{neo-2024-interpreting-context-look-ups} \\
        & & \citet{choi-2024-scaling-automatic-neuron-description} \\
        & & \citet{rajamanoharan-2024-jumping-ahead} \\
        & & \citet{chen-2025-knowledge-microscope} \\
        & & \citet{muhamed-2025-decoding-dark-matter} \\
        & & \citet{movva-2025-hypothesaes} \\
        & & \citet{poche-2025-consim} \\
        \cmidrule(lr){2-3}
        & Detection / Fuzzing & \citet{paulo-2024-automatically-interpreting-millions} \\
        \midrule

        \multirow{10}{*}{\inputtag{} Input-based Evaluation} & \multirow{2}{*}{AUROC} & \citet{heap-2025-saes-can-interpret-randomly} \\
        & & \citet{kopf-2025-prism} \\
        \cmidrule(lr){2-3}
        & \multirow{2}{*}{Mean Activation Difference} & \citet{gur-arieh-2025-output-centric} \\
        & & \citet{kopf-2025-prism} \\
        \cmidrule(lr){2-3}        
        \cmidrule(lr){2-3}
        & Specificity & \citet{templeton-2024-scaling-monosemanticity} \\
        \cmidrule(lr){2-3}
        & Purity, Responsiveness & \citet{puri-2025-fade} \\
        \cmidrule(lr){2-3}
        & F1 & \citet{gao-2025-scaling-evaluating-saes} \\
        \midrule
        
        \multirow{5}{*}{\outputtag{} Output-based Evaluation} & \multirow{3}{*}{Intervention / Steering}
        & \citet{paulo-2024-automatically-interpreting-millions} \\
        & & \citet{gur-arieh-2025-output-centric} \\
        & & \citet{puri-2025-fade} \\
        \cmidrule(lr){2-3}
        & Surprisal Score & \citet{paulo-2024-automatically-interpreting-millions} \\
        \midrule
        
        \multirow{4}{*}{\semtag{} Semantic Similarity} & \multirow{4}{*}{Cosine Similarity} & \citet{lee-oikarinen-2023-importance-prompt-tuning} \\
        & & \citet{paulo-2024-automatically-interpreting-millions} \\
        & & \citet{heap-2025-saes-can-interpret-randomly} \\
        & & \citet{kopf-2025-prism} \\
        \midrule
        
        \multirow{7}{*}{\humantag{} Human Evaluation} & Correctness & \citet{bills-2023-explain-neurons} \\
        & Correctness / Preference & \citet{lee-oikarinen-2023-importance-prompt-tuning} \\
        & Readability & \citet{li-2024-evaluating-readability} \\
        \cmidrule(lr){2-3}
        & \multirow{2}{*}{Mono-/ Polysemanticity Rating} & \citet{rajamanoharan-2024-jumping-ahead} \\
        & & \citet{kopf-2025-prism} \\
        \cmidrule(lr){2-3}
        & Plausibility & \citet{elhelo-geva-2024-maps} \\
        & Faithfulness & \citet{gur-arieh-2025-output-centric} \\
        \bottomrule
    \end{tabular}
    }
    \caption{Concept description \textit{evaluation} techniques categorized by metric, study, and the underlying quality being measured. Metrics are grouped into conceptual families: predictive simulation, input-based evaluation, output-based evaluation, semantic similarity, and human judgment.}
    \label{tab:eval}
    \vspace*{-.1em}
\end{table*}

\section{Evaluating Concept Descriptions}
\label{sec:evaluation}

Evaluating the quality of a concept description is a critical and non-trivial challenge. A good description should be accurate, faithful to the model's internal processing, and understandable to humans. In recent years, the field has developed a diverse toolkit of evaluation techniques, which we categorize into five main families: predictive simulation, input-based evaluation, output-based evaluation, semantic similarity, and human evaluation. Table~\ref{tab:eval} provides a comprehensive overview of these methods, the studies that use them, and the specific quality they aim to measure.

\subsection{\predtag{} Predictive Simulation}
\label{eval:predictive_simulation}
Automated metrics are essential for evaluating descriptions at scale. 
The most widespread automated evaluation paradigm tests a description's predictive power: how well can it be used to simulate the feature's behavior? The canonical example is the simulator correlation method from \citet{bills-2023-explain-neurons}. Here, a ``simulator'' LLM (e.g., GPT-4) is given a feature's description and a text sample, and it must predict the feature's activation value for each token. The quality of the description is determined by the correlation between the simulated and the true activations.
The simulator correlation scores in \citet{bills-2023-explain-neurons} are generally very low (only 1,000 out of 307,200 neurons score at least $0.8$, on a $[0.0, 1.0]$ scale). Adjusting the activations to a $[0, 10]$ scale creates a further loss of precision and nuance.
While widely adopted (Table~\ref{tab:eval}), this method has a key limitation: its reliance on another opaque LLM introduces a layer of confounding abstraction, making it difficult to know if a high score reflects a good description or simply the simulator's own pattern-matching prowess.
Variations on this theme frame the task as classification instead of regression. For instance, the Detection and Fuzzing metrics \cite{paulo-2024-automatically-interpreting-millions} ask a simulator to make a binary decision about whether a given text would activate the feature described. 
Similarly, Automated Simulatability \cite{poche-2025-consim} tests if a description allows a simulator to predict the final output of the explained model, providing a more holistic measure of predictive utility.

\subsection{\inputtag{} Input-based Evaluation}
\label{eval:input-based}
A second family of metrics, building on the framework proposed by \citet{huang-2023-rigorously-assessing-neuron-nles}, evaluates how accurately a description characterizes the inputs a feature activates on. 
The goal is to measure the description's purity (it only covers things the feature responds to) and coverage (it covers everything the feature responds to). 
Metrics like Specificity \cite{templeton-2024-scaling-monosemanticity} and Purity \cite{puri-2025-fade} quantify how selectively a feature fires for inputs matching its description versus random inputs.
These approaches are often contrastive, testing a feature's response to on-concept examples versus ``distractor'' examples \cite{mcgrath-2024-mapping-latent-space}.
This property is often measured with standard classification metrics like AUROC, which assess how well a feature's activation score separates concept-positive from concept-negative examples \cite{heap-2025-saes-can-interpret-randomly, kopf-2025-prism}. Other metrics assess the mean activation difference between on-concept and off-concept inputs, regarding a description as better when activating on-concept examples have higher mean activations than off-concept examples \cite{gur-arieh-2025-output-centric,kopf-2025-prism}.

\subsection{\outputtag{} Output-based Evaluation}
\label{eval:output-based}
The most rigorous metrics test a description's causal faithfulness: does it correctly predict the feature's effect on the model's output? These methods often rely on interventions or steering \cite{paulo-2024-automatically-interpreting-millions,gur-arieh-2025-output-centric,puri-2025-fade}. For example, an evaluator might artificially activate a feature and check if the model's output distribution shifts in the way the description predicts (e.g., making a specific word more likely).
The framework by \citet{paulo-2024-automatically-interpreting-millions} also includes causal metrics, such as direct Intervention Scoring and Surprisal Scoring. The latter measures whether providing the description reduces the model's loss on relevant inputs, offering a proxy for causal impact. These causal evaluations are crucial for verifying that a description is not merely correlational but reflects the feature's actual function. 

\subsection{\semtag{} Semantic Similarity}
\label{eval:semantic_similarity}
When ground-truth concepts are known or can easily be labeled, a straightforward automated evaluation is to measure the semantic similarity between the generated description and the ground-truth label. This is typically implemented by embedding both the generated text and the ground-truth concept name using a sentence embedding model \cite{muennighoff-2023-mteb} and then calculating their cosine similarity. A high score indicates that the generated description is semantically aligned with the expected concept.
This method is used by several studies as a sanity check or for evaluating performance on features with known correlates \cite{paulo-2024-automatically-interpreting-millions, heap-2025-saes-can-interpret-randomly}. Semantic similarity can also be measured in the context of polysemanticity, when multiple descriptions per feature are available \cite{kopf-2025-prism}. In this setting, the reference is the set of feature descriptions, which are descriptive annotations rather than ground-truth labels, and their similarity is measured by comparing them to one another.

\subsection{\humantag{} Human Evaluation}
\label{eval:human}
Ultimately, concept descriptions are for humans, making human judgment essential for validating the meaningfulness of automated metrics. We identify several distinct roles for human evaluation:

\begin{itemize}
    \item \textbf{Accuracy and Plausibility:} The most common use is asking humans to rate if a description is a correct and plausible summary of what a feature does, by showing them high-activating examples \cite{bills-2023-explain-neurons, elhelo-geva-2024-maps}.
    
    \item \textbf{Readability and Clarity:} A description can be accurate but incomprehensible. Works like \citet{li-2024-evaluating-readability} focus specifically on evaluating the linguistic quality and clarity of the generated text.
    
    \item \textbf{Ground-Truth Annotation:} Instead of validating a generated description, humans can be used to create the ground truth itself, e.g., \citet{kopf-2025-prism} employ human annotators to label polysemantic neurons with multiple concepts and apply similarity measures.
    
    \item \textbf{Faithfulness and Usefulness:} Humans can also validate causal claims. \citet{gur-arieh-2025-output-centric} ask humans to assess if a feature's effect on model outputs aligns with its description, providing a human-centric measure of faithfulness.
\end{itemize}

Despite its importance, human evaluation is less common in the mechanistic interpretability community compared to related NLP subfields \cite{geva-2022-ffn-promoting-concepts,simhi-markovitch-2023-interpreting-embedding-spaces}. We posit this is due to two factors: (1) the inherent difficulty, cost, and ambiguity in labeling what a polysemantic component ``does'', and (2) the community's traditional focus on expert-centric, debugging-oriented explanations over layperson-understandable ones \cite{saphra-wiegreffe-2024-mechanistic}.

\section{Findings}
\label{sec:findings}

Our survey of the landscape of concept descriptions reveals several key trends. First, the recognition that neurons can be polysemantic \cite{elhage-2022-superposition} has notable effects on methodological choices. It has driven the widespread adoption of model abstractions like SAEs \cite{bricken-2023-towards-monosemanticity,cunningham-2024-saes-find-highly-interpretable-features}, which aim to provide more monosemantic interpretive units than the neurons themselves. This insight has also inspired the development of new frameworks that can capture multiple concepts per feature, such as \citet{kopf-2025-prism}.

Second, the evaluation of these descriptions is maturing, moving beyond simple correlation scores. The community is developing multi-faceted evaluation metrics to assess descriptions from different angles: their predictive power via simulation (\S \ref{eval:predictive_simulation}), their accuracy at capturing activating inputs (\S \ref{eval:input-based}), and their causal faithfulness to the model's behavior (\S \ref{eval:output-based}). Frameworks that combine these perspectives are becoming the new standard \cite{gur-arieh-2025-output-centric,puri-2025-fade}.

\section{Recommendations for Future Work}
\label{sec:future-work}

\paragraph{From Components to Circuits}
The current focus on describing individual neurons or features is a necessary first step, but true understanding requires knowing how these parts compose. Future work should focus on scaling automated description methods to circuits. While circuit discovery is an active research area \cite{ameisen-2025-circuit-tracing, marks2025sparse}, the next frontier is to generate a natural language description for an entire computational subgraph, explaining how multiple features interact to implement a more complex function.

\paragraph{Scaling to New Data Domains and Modalities}
Most concept description methods have focused on models trained on general web text. A significant opportunity lies in applying these methods to specialized domains. Describing the features of models trained on legal text, medical data, or source code could yield domain-specific insights. Similarly, as models become increasingly multilingual and multimodal, future work should explore how concepts are represented across languages and whether unified descriptions can be found for features that respond to multiple modalities.

\paragraph{Analyzing the Interpreters Themselves}
As we increasingly rely on LLMs to generate explanations, we must critically analyze the ``interpreters'' themselves. What types of concepts are LLM-based describers biased towards? Do they tend to produce simple, atomic descriptions (e.g., \textit{names of cities}) while missing more abstract or relational ones (e.g., \textit{syntactic subject-verb agreement})? Future work should include a meta-analysis of the generated descriptions to understand their linguistic properties, potential biases, and conceptual limitations, ensuring that our window into one model is not distorted by the lens of another.

\paragraph{A Finer-Grained View of Polysemanticity}
SAEs have become the default solution to polysemanticity, but they are not a silver bullet. Future research should explore more nuanced models of feature activation. For instance, rather than assuming a feature is simply ``on'' or ``off'', its activation level may matter; a feature might represent different concepts at different intensities or in different ranges \cite{haider-2025-neurons-speak-in-ranges}. This calls for description methods that can capture this fine-grained detail, such as \textsc{PRISM} \cite{kopf-2025-prism}, to provide a more complete picture of a feature's function.

\paragraph{Rigorous Causal Evaluation}
While evaluation methods are improving, most automated metrics remain correlational. The field must continue to push towards more rigorous tests of faithfulness. This includes scaling up intervention-based methods that test the causal effects of features on model outputs \cite{paulo-2024-automatically-interpreting-millions} and developing ``stress tests'' that assess whether descriptions hold up under adversarial or out-of-distribution contexts. Critiques of the steerability of SAE features suggest that their causal impact is not yet fully understood, making this a critical area for future work \cite{wu-2025-axbench}.

\paragraph{Standardized Benchmarks for Evaluation}
Finally, a significant accelerator for progress in concept description methods would be the development of standardized benchmarks inspired by the Mechanistic Interpretability Benchmark of \citet{mueller-2025-mib}. Future work could build on this by creating benchmarks designed specifically to evaluate the quality of natural language descriptions. Such a benchmark could include a suite of components with agreed-upon ``gold'' descriptions, allowing for a more systematic comparison of different description and evaluation methods.

\section{Conclusion}
In conclusion, the use of LLMs to generate concept descriptions for model components and abstractions represents a notable step towards demystifying the internal workings of LLMs.
Our survey has charted a clear trajectory in this emerging field: from early attempts to label individual, often polysemantic, neurons to higher-level abstractions like SAEs and circuits.
Concurrently, the methods for evaluating these descriptions have matured from simple predictive correlations to a multi-faceted toolkit encompassing input-based purity, causal faithfulness, and nuanced human judgment.

For practitioners, this field offers a powerful new lens for model analysis, debugging, and auditing. However, the path from a generated description to a verifiable causal mechanism is not yet fully paved. The high computational costs and the ongoing debates around the faithfulness of these descriptions mean they should be applied with a critical eye. By continuing to refine the methods for generating these descriptions and, crucially, developing more rigorous and standardized benchmarks for their evaluation, the research community can forge these techniques into indispensable tools for building more robust, transparent, and trustworthy NLP systems.

\section*{Limitations}
For scoping this survey, we limit our focus to natural language concept descriptions for internal model components and abstractions. We intentionally exclude other important families of interpretability work, such as purely mathematical analyses of model properties, feature visualization techniques that do not produce textual output, and methods focused on explaining final predictions rather than internal functions.

We also note that the computational and financial costs associated with the methods surveyed are substantial. Training high-quality SAEs requires plenty of GPU resources, and the subsequent steps of generating descriptions and performing automated evaluations often rely on expensive API calls to proprietary models. These costs currently pose a barrier to wider adoption and reproducibility, particularly in academic settings.

The vast majority of the research surveyed here focuses on English-language models and text corpora. The extent to which these methods and the concepts they uncover generalize to other languages remains a largely open and important question for future investigation.

\section*{Acknowledgements}
We thank the anonymous reviewers at the BlackboxNLP Workshop for their feedback.
We acknowledge support by the Federal Ministry of Research, Technology and Space (BMFTR) for \hbox{BIFOLD} (ref. 01IS18037A) and news-polygraph (ref. 03RU2U151C).

\bibliography{custom}

\end{document}